\def\year{2017}\relax
\documentclass[letterpaper]{article}
\usepackage{aaai17}
\usepackage{times}
\usepackage{helvet}
\usepackage{courier}
\usepackage{amsmath}
\usepackage{amsfonts}
\usepackage{graphicx}
\usepackage{pdfpages}

\usepackage{booktabs}
\usepackage{url}
\usepackage{caption}
\usepackage{subcaption}

\usepackage{xcolor}
\definecolor{ubuntu}{HTML}{E95420}
\definecolor{ubuntu_v}{HTML}{772953}

\newcommand{\ubuntuworld}{\texttt{\textcolor{ubuntu_v}{UbuntuWorld 1.0 LTS}}}

\DeclareMathOperator*{\argmax}{arg\,max}

\usepackage{etoolbox}
\makeatletter
\patchcmd{\@verbatim}
  {\verbatim@font}
  {\verbatim@font\footnotesize}
  {}{}
\makeatother

\begin{document}


%

\title{\ubuntuworld~~-~~A Platform for Automated Problem\\ Solving \& Troubleshooting in the Ubuntu OS}
\author{
Tathagata Chakraborti$^1$ \and Kartik Talamadupula$^2$ \and Kshitij P. Fadnis$^2$\\
{\bf \Large Murray Campbell$^2$ \and Subbarao Kambhampati$^1$}\\
$^1$Department of Computer Science, Arizona State University, Tempe, AZ 85281, USA\\ 
{\tt\small \{ tchakra2, rao \} @ asu.edu }\\
$^2$IBM T. J. Watson Research Center, Yorktown Heights, NY 10598, USA\\
{\tt \small \{ krtalamad, kpfadnis, mcam \} @ us.ibm.com}
}
\maketitle

\begin{abstract}
In this paper we present \ubuntuworld~- a platform for developing automated technical support agents in the Ubuntu operating system. Specifically, we propose to use the Bash terminal as a simulator of the Ubuntu environment for a learning-based agent, and demonstrate the usefulness of adopting reinforcement learning (RL) techniques for basic problem solving and troubleshooting in this environment. We provide a plug-and-play interface to the simulator 
as a python package 
where different types of agents 
 can be plugged in and evaluated, and provide pathways for integrating data from online support forums like \texttt{Ask Ubuntu} into an automated agent's learning process. Finally, we show that the use of this data significantly improves the agent's learning efficiency.  We believe that this platform can be adopted as a real-world test bed for research on automated technical support.
\end{abstract}

\noindent Building effective conversational agents has long been the holy grail of Artificial Intelligence~\cite{turing1950computing}. Research in this direction has, however, largely recognized that different modes of conversation require widely different capabilities from an automated agent, depending on the particular context of the interaction; the focus has thus been on approaches targeted at specific applications. For example, conversational agents in the form of chat bots are required to be more creative, responsive and human-like; while for automation in the context of customer service, qualities like precision and brevity are more relevant. Indeed, human agents while providing customer support make a conscious effort to be as structured as possible in their interactions with the user. For successful automation in this particular mode of dialog (that we refer to as end-to-end goal-directed conversational systems or \texttt{e2eGCS}) we identify the following typical characteristics - 

\begin{itemize}
\item[-] \emph{End-to-end.} This is the ability of the agent to build and operate on knowledge directly from raw inputs as available from the world, and to generate the desired behavior.
\item[-] \emph{Goal-directed.} The interactions in these settings are targeted at achieving specific goals, i.e. to solve a particular problem or reach a desired state.
\item[-] \emph{General purpose.} It is infeasible to build fundamentally different support agents for every possible environment, and hence there must be a learning component to the agent that facilitates automated building of domain knowledge.
\item[-] \emph{Adaptive.} An agent must learn to adapt to its experience and update its knowledge, and this further underlines the importance of an agent's capability to learn.
\item[-] \emph{Integrated.} Finally, the agent must be able to interact with the human in the loop and integrate (and subsequently learn from) human intelligence in order to solve a wide variety of problems effectively.
\end{itemize}

\noindent One of the canonical examples of such systems is \emph{technical support}. As in the case of customer service in general, automation for technical support requires an agent ascribing to the \texttt{e2eGCS} paradigm to be able to:
\begin{itemize}
\item \textbf{learn} a model or understanding of its environment automatically by means of experience, data and exploration;
\item \textbf{evaluate} its knowledge given a context, and learn to sense for more information to solve a given problem; and
%
%
%
\item \textbf{interact} with the customer, maybe in multiple turns, in a natural fashion to solve a given problem effectively.
\end{itemize}

\noindent In this paper \emph{we specifically address the learning problem}, and make a first attempt to lay a pathway towards achieving fully fleshed-out \texttt{e2eGCS} of the future. 
Technical support is a particular instance of customer service that deals with problems related to the operation of a specific piece of technology, which means there often exists an underlying (albeit unspecified) model to the operation of such a system, and the model learning proposition becomes especially attractive in this context. 
However, the critical problem here is that the engineers who build the technology, the people who use it, and the ones who provide support for it are often distinct from each other. 
One solution then would be to make the architects of the system also build the support engine following the same software specifications; this quickly becomes intractable (and might well require its own support!). 
A more worthwhile alternative is to learn this model automatically. Such an approach, while being considerably simpler to follow, is also likely to be more effective in capturing domain knowledge and providing directed personalized support. 

\begin{figure}[btp!]
\includegraphics[width=\columnwidth]{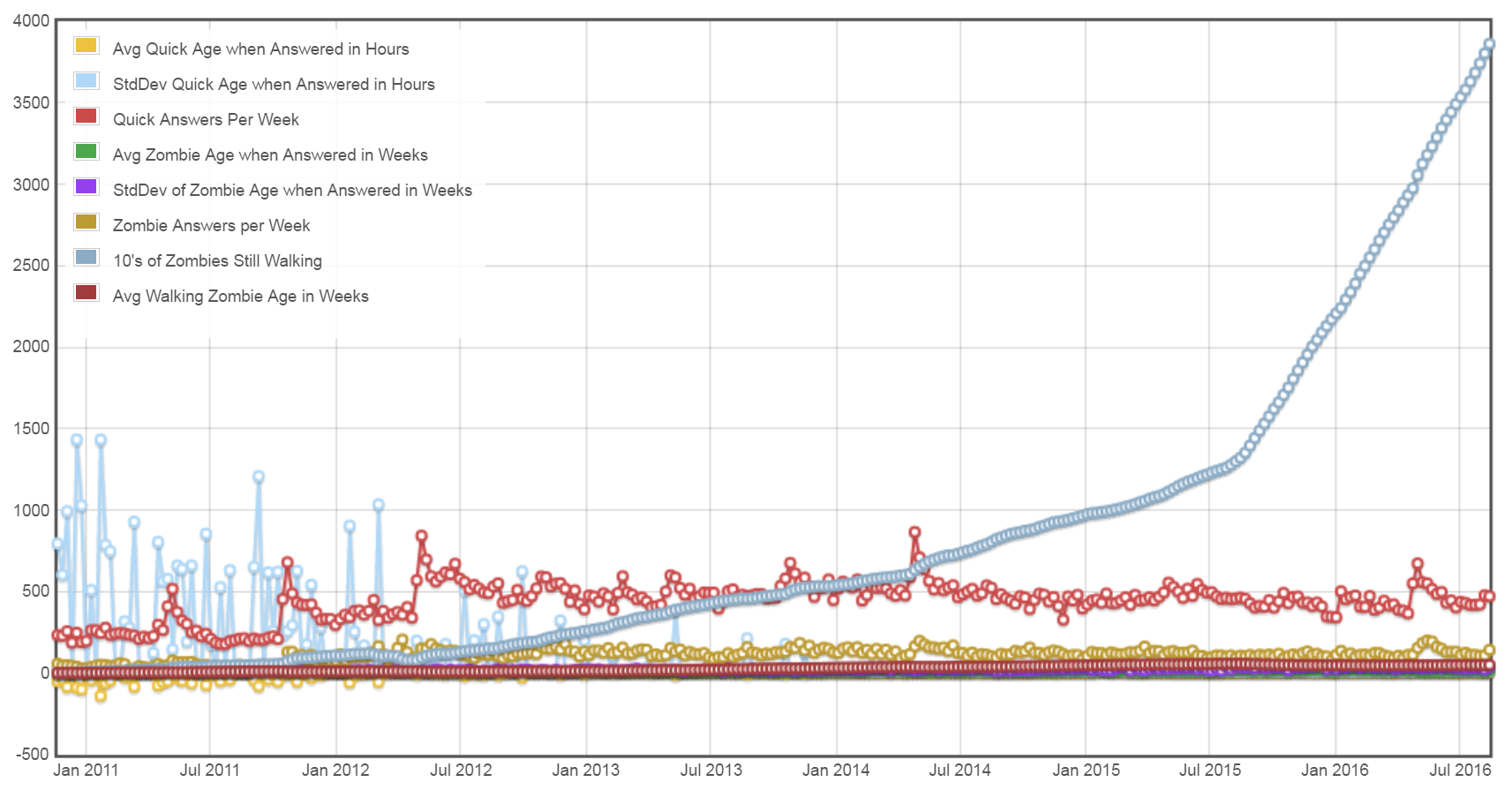}
\caption{Graph (\url{http://bit.ly/2blmZk1}) 
showing the number of \emph{zombie posts} from 01/2011 to 07/2016. These are the posts that have remained unanswered for more than 72 hours, and their number is growing exponentially.}
\label{zombie}
\end{figure}

The specific domain we look at in this work is technical support in the Ubuntu operating system. This is undoubtedly a real-world environment where support is extremely sought after. Indeed there is a thriving community on the online Ubuntu help forum \texttt{Ask Ubuntu}, a question and answer site for Ubuntu users and developers hosted on the Stack Exchange network of Q\&A sites. \texttt{Ask Ubuntu} currently boasts more than 370k registered users and 238k questions asked till date, and ranks third overall in the family of 158 Stack Exchange communities in terms of traffic or number of users (as of August 2016). A closer look however reveals that this rank is not an indicator of the quality of support. In terms of percentage of questions actually answered~\cite{percentage}, \texttt{Ask Ubuntu} operates at a lowly rate of $65\%$, ranking just five places off the bottom of the list. Further, as shown in Figure~\ref{zombie}, the number of posts that go unanswered is exploding in recent times \cite{outofcontrol}. While there are many causes that may have led to these dire circumstances, some of which we discuss below, one thing is quite certain - \emph{Ubuntu needs support, and there isn't enough of it out there}.

\subsubsection{Motivation}

\texttt{Ask Ubuntu}'s afflictions may be largely attributed to the following main causes - 

\begin{itemize}
\item[1.] New users clogging up the system with simple problems that experienced users do not care to respond to.
\item[2.] Duplicate questions, due to large numbers of users who do not bother to look up existing solutions before posting. 
\item[3.] An unhealthy newcomer to expert ratio in the community as a result of Ubuntu's rapidly growing popularity.
\item[4.] The continuous roll out of new software/OS versions and corresponding problems with dependencies.
\item[5.] Incompletely specified problems, including insufficient state information and error logs leaving members of the community little to work with.
\end{itemize}

We claim here that a large number of these problems can readily be solved through automation. While it may not be reasonable to expect an automated agent to learn the most nuanced details of the Ubuntu OS and solve niche issues that the experts on \texttt{Ask Ubuntu} are more capable of addressing, the large majority of problems faced by users on the forum are readily addressable. These are either (1) simple problems faced by newbies that may be directly solved from the documentation, whose solutions can be learned from exploration in the terminal; or (2) duplicates of existing issues which may have already been solved, whose solutions may be retrieved using relevant data from \texttt{Ask Ubuntu}. The learning approach then also indirectly addresses issues (3) by freeing up (and in turn tapping into) support from \texttt{Ask Ubuntu}; and (4, 5) since the domain knowledge built up over time as well as local state information sensed by the integrated support agent may be useful in providing more directed and personalized support.

\begin{figure}[tbp!]
\includegraphics[width=\columnwidth]{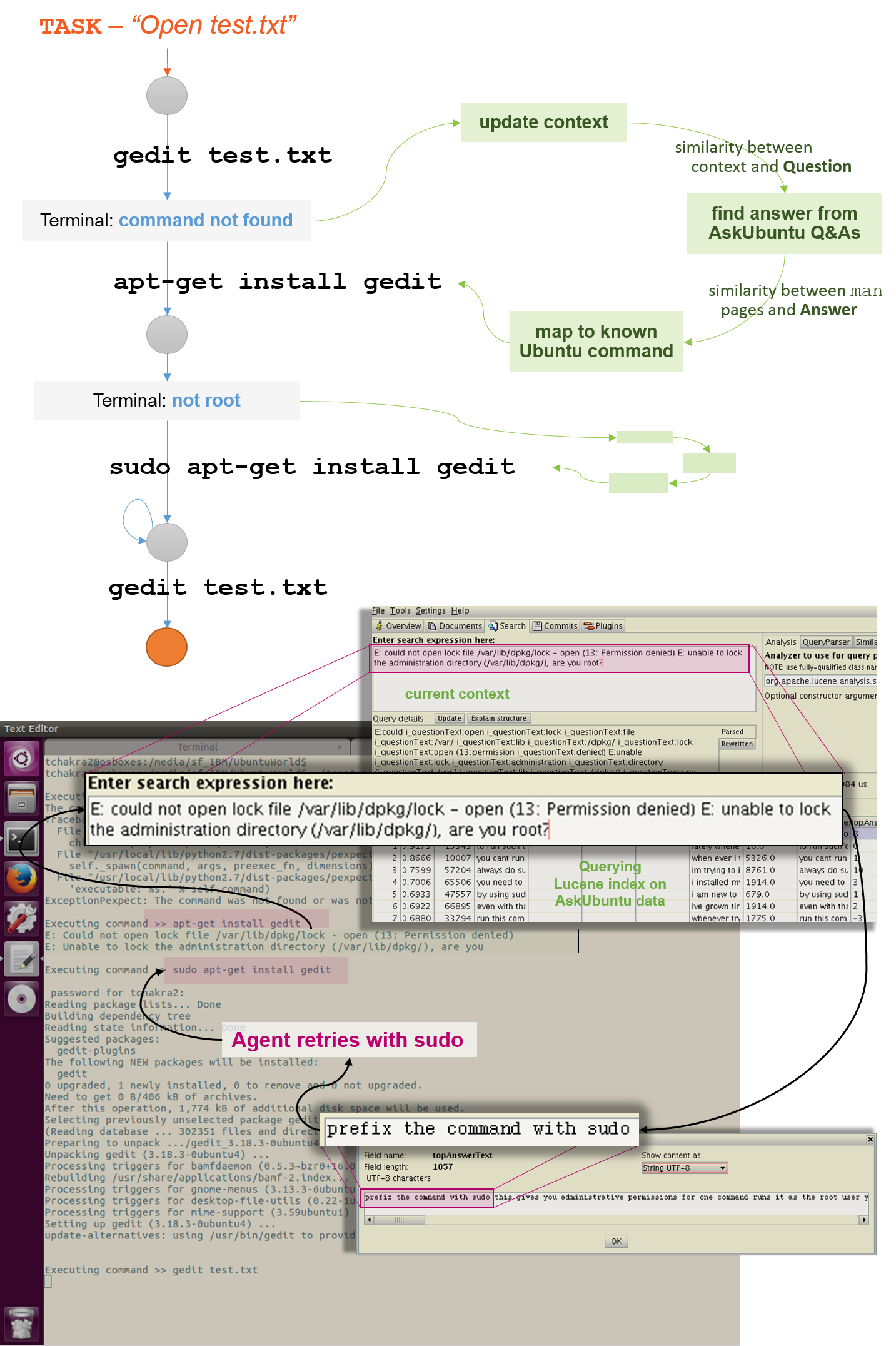}
\caption{Use case - querying \texttt{Ask Ubuntu} for guidance.}
\label{use_case}
\end{figure}

\begin{figure*}[tbp!]
\includegraphics[width=\textwidth]{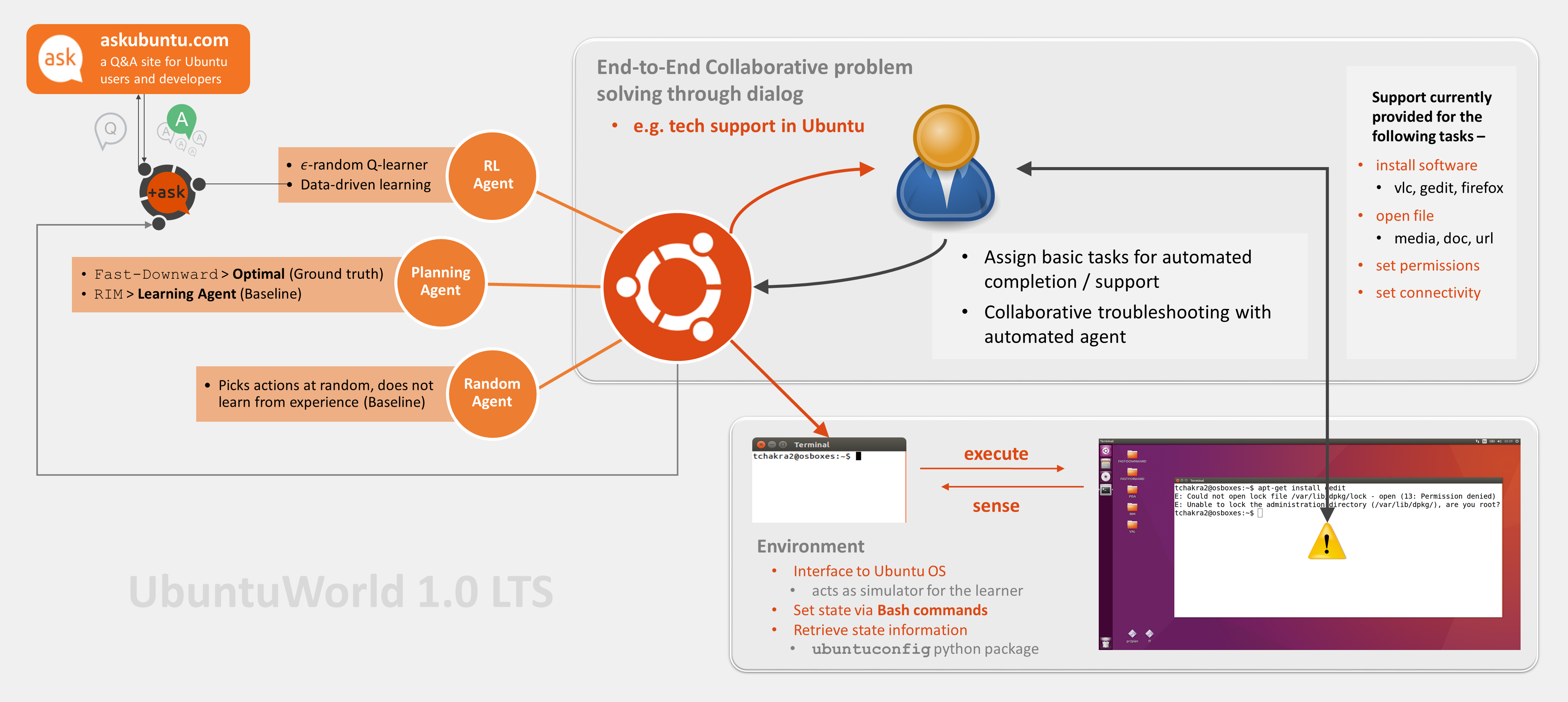}
\caption{A schematic representation of the \ubuntuworld~prototype.}
\label{schematic}
\end{figure*}

Figure~\ref{use_case} provides an illustration of the kind of tasks we are interested in. Consider the simple task of opening a text file. It can be achieved in a single step using \texttt{gedit}, or it can be arbitrarily harder depending on the actual state of the system - the agent might need to install \texttt{gedit} if it is not available, and it may need to access the internet and gain \texttt{sudo} permissions to install \texttt{gedit}. This is represented in the top part of the figure. We want our agent to learn these workflows and dependencies on its own, by exploring the Bash environment. Thus when an error comes up, for e.g. regarding administrative rights, the agent knows it needs to execute the \texttt{sudo} command. 
Of course, this is one of a very large number of traces that the agent will need to explore before it converges on the correct choices, and this is hard for a setting such as Ubuntu due to the large number of actions that an agent can possibly perform at any given state. Perhaps ironically, we turn to \texttt{Ask Ubuntu} itself in order to make the learning agent's life a little easier in this regard. 

As we noted before, users on \texttt{Ask Ubuntu} have been solving these problems for a long time, and their solutions can in fact be used to provide valuable guidance to a learning agent, as shown in 
in Figure \ref{use_case} - here, the terminal has produced a \emph{``Permission denied''} error in response to a call to an \texttt{apt-get} command without a \texttt{sudo} prefix. The agent in response queries \texttt{Ask Ubuntu} with the text of the error by using a pre-built \texttt{Lucene} \cite{lucene} reverse index, and finds answers such as \emph{``prefix the command with sudo''} that can be used to inform its own deliberative process. The recommended action -- in this case, \texttt{sudo} -- is then extracted by (TF-IDF based) similarity matching between the top retrieved answers and the description of the actions (we used linux \texttt{man} pages for this) in our domain.
Thus, in addition to wanting our learning agent to explore and build a model of Ubuntu, we also want to make sure that the exploration is smart given the abundance of data already available on troubleshooting in Ubuntu. 
We refer to this as \emph{data-driven reinforcement learning}, and we expand on this idea through the rest of the paper.

\subsubsection{Related Work} 
Bringing AI techniques - particularly reasoning and decision-making - to the problem of automated software agents has a rich history within the automated planning community. Of particular importance is the work on building softbots for UNIX \cite{etzioni1993building,Etzioni:1994:SII:176789.176797}, which is the most comprehensive previous study on this problem. Indeed, as we introduced earlier, many of the issues that are cataloged in that work remain of importance today. The current work builds upon the work of Etzioni et al., particularly their focus on a goal-oriented approach to the problem~\cite{etzioni1993agents}; however, it goes beyond that work (and related approaches \cite{Petrick:2007}) in actually realizing a \emph{learning-based} agent framework 
for the Ubuntu technical support domain. Succinctly, we seek to automate in the largest possible degree the promise of the softbot approach by: (1) exploiting the Bash shell as a robust simulator for learning agents to explore the world; and (2) using the large amounts of data generated by human experts on the internet.

On the learning side, Branavan et al.'s work on using reinforcement learning (RL) to map natural language instructions to sequences of executable actions~\cite{branavan} explores a similar problem setting in a Windows OS domain. However,  that work focuses on the application of RL techniques to the language processing problem, and on mapping text instructions to executable actions. In contrast, our work focuses on learning task-oriented models for solving the \texttt{e2eGCS} problem. Thus the most relevant prior explorations into this area are complementary to our work in different ways; while the softbot work lays the groundwork for our architecture, Branavan et al.'s work provides a report on using learning on a related but different problem. 


\subsubsection{Contributions} 
The contributions of the paper are - 
\begin{itemize}
\item We provide a platform \ubuntuworld~based on the Ubuntu OS, and its interface to the Bash terminal, where different types of agents can be plugged in and evaluated. This can be adopted as a valuable real-world test bed for research in automated technical support.
\item We propose data-driven RL as a viable solution to the model learning problem for automated technical support in Ubuntu, by utilizing human intelligence -- in the form of data from online technical support forums like \texttt{Ask Ubuntu} -- to aid the traditional RL process.
\end{itemize}


\section{\ubuntuworld}
\label{system}

The main components of the proposed system (Figure~\ref{schematic}) are the agents, the environment, and the user. 
As mentioned previously, the environment is the Ubuntu operating system. 
The user and the agents are the two main actors in the setting -- they interact with the environment, or with each other, in different capacities to perform tasks.

\subsection{The Agent Ecosystem} 

Though the user interacts with a generic, fixed agent interface, the internal nature of that agent could be one of several types depending on the type of technology used:

\subsubsection{Random Agent} The Random Agent does not have any learning component: it performs actions at random till it achieves its goals. This is used as a baseline to evaluate how difficult the planning problems are, and how much a learning agent can gain in terms of performance.

The next two agents we describe make use of a state-based representation of the Ubuntu environment in order to make more informed choices about the next action to execute. 
An example of such a state is shown below. 
In the current implementation, the relevant variables and predicates need to be provided by the software developer for both approaches.

\begin{verbatim}
internet-on : True
sudo-on : False
installed gedit : False
installed firefox : True
installed vlc : False
open gedit file : False
open firefox file : False
open vlc file : False
\end{verbatim}

\subsubsection{Planning Agent} The Planning Agent uses PDDL models~\cite{pddl} of 
the domain to compute plans. It is integrated with the \texttt{Fast-Downward} planner \cite{helmert} that can be used to produce the optimal plan given a problem and domain description. The problem description is built on the fly given the current state being sensed by the agent and the target (partial) goal state. An excerpt of the domain is shown below (a link to the entire domain file and a sample problem for a simple \emph{``open file''} task are available at \url{http://bit.ly/2c8kJ4Q} and \url{http://bit.ly/2clwwKI} respectively):

\begin{verbatim}
(:action AptGet_True
:parameters	    (?s - software)
:precondition   (and (sudo-on) 
                     (internet-on))
:effect         (and (installed ?s)))

(:action VLC_True
:parameters     (?o - item)
:precondition   (and (not (sudo-on)) 
   	                 (installed vlc))
:effect         (and (open vlc ?o)))
\end{verbatim}

\noindent The domain itself may either be hand-coded from the software developer's knowledge
, or \emph{learned} from execution traces \cite{rim}. The former can serve as the ground truth for evaluating the performance of various agents, while the latter provides a valuable baseline to compare against the other learning agents. 

\subsubsection{RL Agent} The reinforcement learning (RL) paradigm involves learning policies or models of the environment by acting and learning from experiences in the world and associated feedback. One of the standard forms of RL is Q-learning~\cite{sutton}, where an agent learns a function $Q:S \times A \rightarrow \mathbb{R}$ that maps state-action pairs to real values that signify the usefulness or utility of doing action $a \in A$ in state $s \in S$. The learning step is the well-known Bellman update when a transition from state $s$ to $s'$ is observed due to an action $a$, and a reward $R:S \times A \times S \rightarrow \mathbb{R}$ is received -
{\small
\begin{align*}
Q(s, a) \leftarrow (1-\alpha)Q(s, a) + \alpha\{R(s,a,s') + \gamma\max_{a\in A}Q(s', a)\}
\end{align*}
}%
Here, $\alpha$ is the learning rate, and $\gamma$ is the discount factor. 
During the learning phase, the agent does an exploration-exploitation trade-off by picking an action $a$ given the current state $s$ (represented by $a|s$) based on several intentions given probability thresholds $\epsilon, \beta$, and $(1 - \epsilon - \beta)$, as shown below. This forms the core of what we refer to as the \emph{``data-driven" $\epsilon$-random Q-learning agent}.
%
%
{
\begin{align}
\label{eqn1}
a|s & \leftarrow \argmax_{a\in A}Q(s, a)
\end{align}
}
\noindent \emph{- Exploitation of the learned representation} - This option (Equation \ref{eqn1}) allows the RL agent to pick the action with the maximum Q-value in the current state. In the current implementation, the system employs a tabular Q-learning approach where the state variables are the Boolean predicates (in the domain as well as those that appear in the goal) from the reference planning model shown above. Note that the state representation 
integrates both the goal information and the current value of the state variables, in order to ensure that the agent learns goal-directed policies. 
{
\begin{align}
\label{eqn2}
a|s & \sim \mathcal{U}(A)
\end{align}
}
\emph{- Random Exploration} - This is done by choosing the next action randomly with probability $\epsilon$ (Equation \ref{eqn2}). This is the standard $\epsilon$-random Q-learning agent. 
{
\begin{align}
\label{eqn3}
a|s & \leftarrow \argmax_{a\in A} \mathcal{D}_a \cap \texttt{AskUbuntu}^+(\mathcal{F}_a)
\end{align}
}
\emph{- Exploration by querying the \texttt{Ask Ubuntu} data}
- Here the agent explores by choosing an action $a$ that maximizes the similarity between the action documentation $\mathcal{D}_a$ and the relevant questions and their solutions in the forum posts, in order to pick the next best action (Equation \ref{eqn3}). The action documentation $\mathcal{D}_a$ of an action $a$ in our implementation is the entire content of the \texttt{man} page associated with that action. The relevant questions and answers are retrieved by querying \texttt{Ask Ubuntu} with the footprint $\mathcal{F}_a$ of action $a$. The action footprint in our case is the text output on the terminal as a result of executing that action.
The accepted answers to the top 5 posts are then used to query the man page descriptions of the Ubuntu commands available to the agent in order to determine the set of relevant actions to perform next. The action that produced the maximum similarity was selected for execution. 

The parameter $\beta$ was varied according to a damped sine function to alternate between the $\epsilon$-random exploration mentioned above, and the data driven exploration described here. The intuition behind this is to alternate between the suggestions from \texttt{Ask Ubuntu} and the agent's own exploration function alternatively early on, and then gradually fall back on exploitation later. This ensures that the agent can utilize good suggestions from the forum to guide the initial learning process, but at the same time does not get stuck with bad suggestions that may creep in either due to noise in the data or during the retrieval and similarity matching step. The results of this scheme are detailed in the evaluation section.

For our environment, the reward function is defined as follows: 
The agent gets a negative reward every time it does an action, so that it learns to prefer shorter policies.
If, however, the state changes due to an action, the amount of negative reward is less, since the agent at least tried an action that was applicable in the current state.
Finally, there is a large reward when the agent attains a state that models the goal. 
{\small
\begin{align*}
R(s, a, s') & = -10 \text{ if } s' \not= \bot\\
& \mathrel{+}= 5 \text{ if } s' \not\models s\\
& \mathrel{+}= 100 \text{ if } s' \models \mathcal{G}
\end{align*}
}%
Note that this definition of the reward function is generic to any environment, and does not preclude added (domain specific) information such as whether an action was successful or not, etc. 
This also means that the learning process will suffer from the same issues that are common in traditional RL techniques with regards to \emph{delayed} rewards, and the many approaches that have been investigated in the contemporary literature to mitigate such problems also apply here.

\subsection{The Environment} 

The Environment in our case is the Ubuntu OS, which both the agent and the user have access to via Bash commands on the terminal. Through the terminal the agent can execute actions, as well as sense the state of different environment variables and the current output on the terminal. The way these interactions are used depends on the specific type of the agent. Currently, the agents only have access to actions whose effects are all reversible, i.e. the \ubuntuworld~environment is currently ergodic.

\subsection{Agent Interactions}
As mentioned previously, both the user and the agent can interact with the environment through the terminal to accomplish specific tasks. The user can also interact with the agent and ask it to complete basic tasks (automated problem solving), as well as invoke the agent in case she encounters an error on the terminal (automated troubleshooting). The agent may, in trying to solve a task, interact with the user in trying to find the correct parameters for an action or ask for more clarifications to solve the task, or even query \texttt{Ask Ubuntu} to search for a possible solution to a problem. 

\subsubsection{Implementation Details}
\label{code}

\begin{figure}[t!]
\centering
\includegraphics[width=0.95\columnwidth]{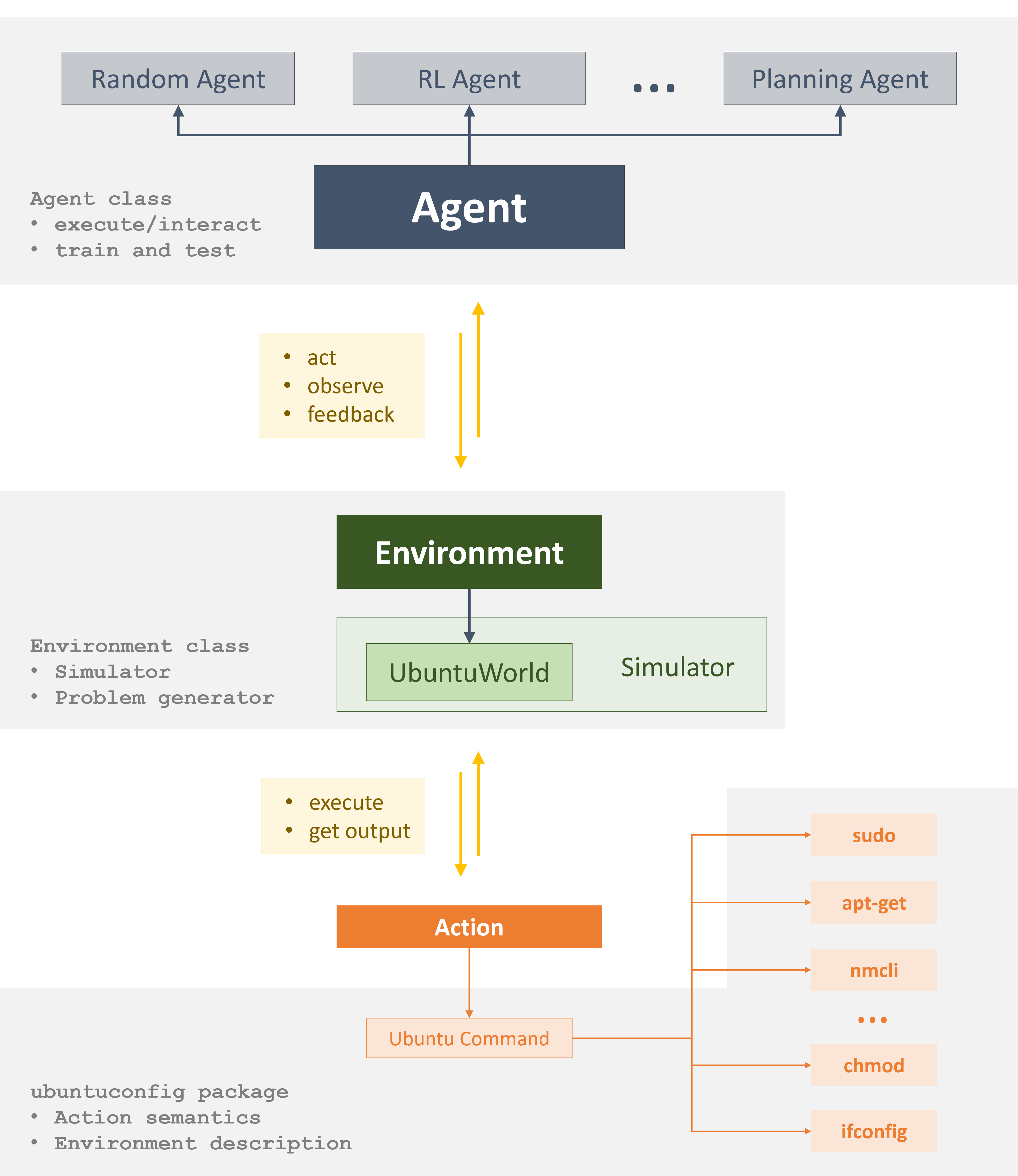}
\caption{Architecture diagram showing different components of the \ubuntuworld~ prototype.}
\label{arch}
\end{figure}

The system architecture has three main components (Figure~\ref{arch}) - the Agent Class, the Environment Class and the \texttt{ubuntuconfig} package.

\subsubsection{\textnormal{\em The Agent}} 

may be asked to solve a task, or train and test on a set of problem instances. The base agent implements the Random Agent, while all the other agents such as the Planning Agent and the RL Agent inherit from it.  
The key difference is (1) how, given a state, the \emph{``get next action''} process is done, e.g. the Random Agent picks the next action at random, the Planning Agent re-plans optimally from the current state and picks the first action from the remaining plan, and the Q-learning RL Agent picks the action that has the maximum Q-value in the current state; and (2) what the Agents do with the result of executing he action, e.g. the Random Agent ignores it, while the learning agents may use it to learn a representation - such as a PDDL domain or a Q-function - of the environment. Finally the Agents also have abilities to take snapshots of themselves and reboot, and display learning curves and progress statistics during training.

\subsubsection{\textnormal{\em The Environment Class}} 

acts as the interface between the Agent Class and the \texttt{ubuntuconfig} package by using generic interaction semantics - the agent can access the Environment by sending an action to it and receiving the output as a result of it, while the specific environment package implements the actual execution pathways and provides the action footprints to the environment. Thus specific agents and environments may be swapped in and out while their interface remains identical.

Thus the two main functionalities of the Environment Class are (1) reading in an environment description, such as from \texttt{ubuntuconfig}, and setting up the environment; and (2) simulating this environment as required by the agents plugged into it. It can also generate training and testing problem instances given the environment description. The UbuntuWorld Class inherits from the basic Environment Class and implements Ubuntu specific methods that can sense and set values of state variables as required. 

Finally, it may not always be a good idea to run the agents on an environment directly - e.g. installing software takes time, and trying to train agents whose potential performances are completely unknown may be a waste of resources. Keeping this in mind, the Environment Class also implements a wrapper that emulates an environment description without running it. This, of course, cannot be done with the full environment, since the model is not known. It can, however, in the \emph{simulation mode} emulate a known part of the environment and help in debugging and setting up (for example) the parameters of the learning agent, etc. 

\subsubsection{\textnormal{\em The~~\texttt{ubuntuconfig}~~Package}} 

contains the description of the UbuntuWorld domain, i.e. the actions available to the agent, state variables that can be sensed, methods to execute each of these actions and parse their outputs, etc.

Each action in the UbuntuWorld environment is implemented as a separate class - the individual classes implement how the interactions with the terminal play out for specific actions or commands, e.g. a permissions (sudo) check followed by a memory usage check for the \texttt{apt-get} command. Each action class comes with methods to execute it, get its output, and optionally check for its success (this is not used in the RL setting since the model is not known). 

The Command Class implements the basic functionalities of all commands, including a generic interaction with the shell with or without invocation with the sudo prefix. Specific action classes inherit from it and implement their own parameters and shell interactions. Apart from the modular and concise nature of the command definitions, making the Ubuntu commands available as separate class objects also leaves the processing at an agent's end as general purpose as possible, with scope for caching and reuse depending on the nature of the agent.
If the commands do not have any unique semantics, then these command classes are generated automatically from a list of the action names and their bindings to specific Bash commands in Ubuntu. Since this is the case most of the time (i.e. most Bash commands do not involve sophisticated interactions with the shell) this alleviates scalability concerns with this particular approach, while at the same time providing surprising flexibility with how the Ubuntu shell may be accessed by automated agents.

\section{Experiments and Looking Forward}
\label{eval}

As a preliminary evaluation of our system, the environment was set up to handle open/close, install/remove, internet access, and root privilege tasks as discussed before in Figure \ref{use_case}. In the following, we will discuss the relative performance of the our data-driven RL agent in the context of these tasks.

\subsubsection{Learning rate.}

\begin{figure}[t!]
    \centering
    \begin{subfigure}[]{\columnwidth}
\includegraphics[width=\columnwidth]{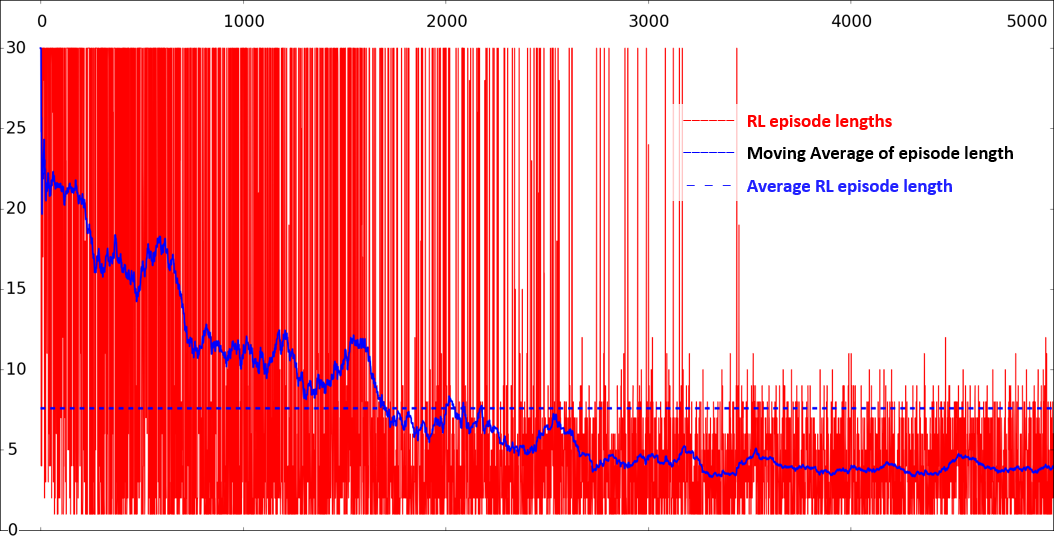}
\caption{Learning performance (episode lengths) of the RL agent in course of training over 1000 problems instances, replayed four additional times. The episodes were terminated after 30 steps. The agent shows clear signs of learning beyond 3000 episodes.}
\label{train}
\vspace{15pt}
    \end{subfigure}
    \begin{subfigure}[]{\columnwidth}
\includegraphics[width=\columnwidth]{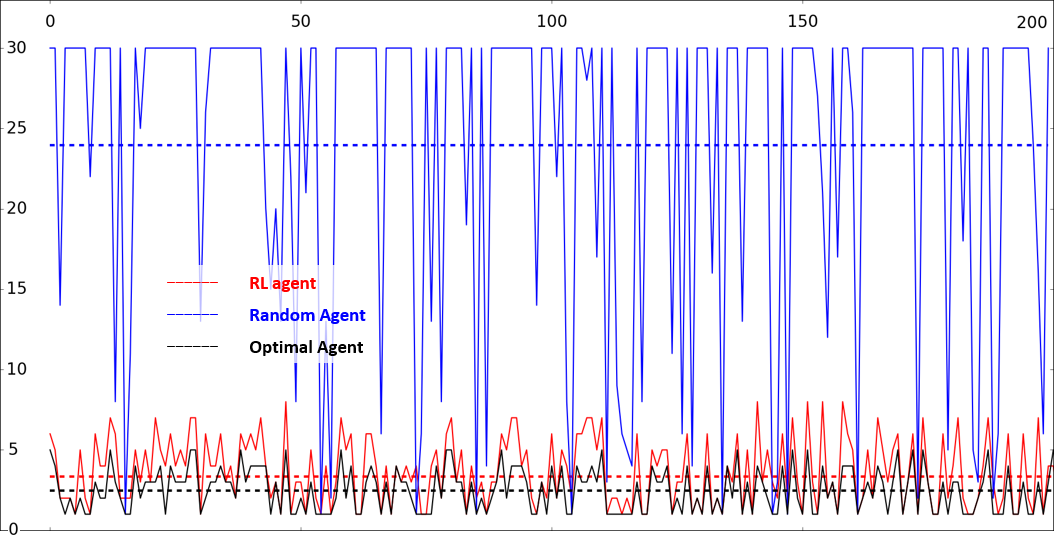}
\caption{Test performance (plan lengths) of the RL agent in 200 randomly generated tasks, against the optimal and the random agents. The performance is close to and mimics closely the optimal plans, while being a significant improvement from the random agent.}
\label{test}
    \end{subfigure}
    \caption{Training and testing performances on simple tasks involving opening files from various start configurations.}
\label{results}
\end{figure}

Figure~\ref{results} shows the performance of a simple $\epsilon$-random Q-learning RL Agent trained on the emulator on simple tasks involving opening files, as described before. We measure the performance of an agent in terms of the lengths of the sequences (plans) required to solve a given problem, and compare these with those of the optimal plan. This optimal length is generated by the Planning Agent using the underlying complete PDDL model (which acts as the ground truth), and the Random Planner (which acts as a simple baseline). Figure \ref{train} shows convergence of the agent beyond around 3000 episodes, as the moving average length settles around the ground truth range (original plans require up to 5 actions). The test performance of the agent is shown in Figure \ref{test}. The agent mimics the optimal plans impressively, and is significantly better than a random agent, signifying that the learning tasks are non-trivial as well as the fact the tasks have been learned effectively by the agent.

\subsubsection{The data-driven agent.}

\begin{figure}[t!]
\centering
\includegraphics[width=\columnwidth]{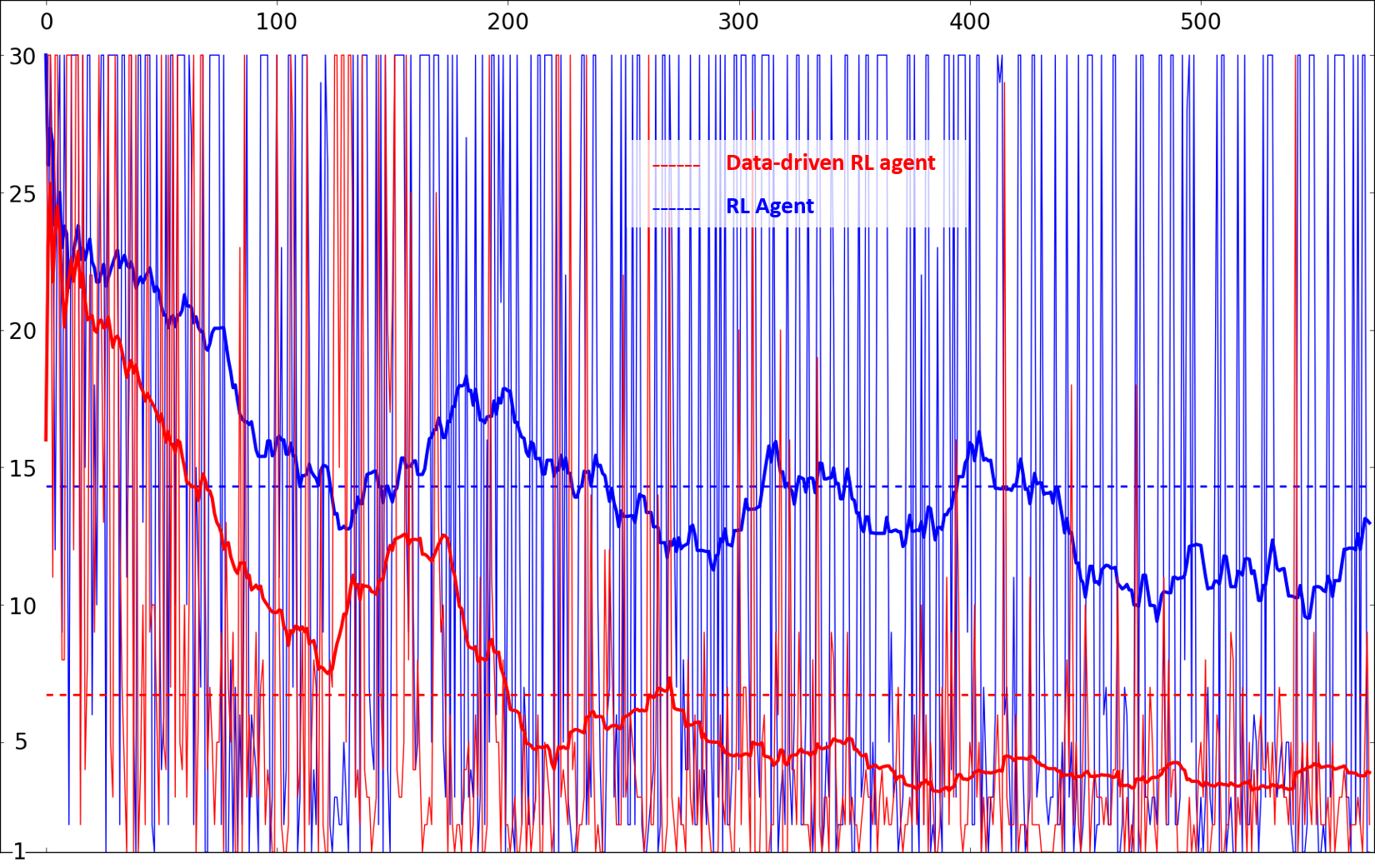}
\caption{Superior learning rate of the data-driven RL Agent underlines the value of leveraging data from online technical support forums like \texttt{Ask Ubuntu} in providing guidance to the model learning process of an automated agent.}
\label{data}
\end{figure}

Figure~\ref{data} shows the relative convergence rates of an $\epsilon$-random RL Agent with and without data support, run on the actual terminal without the emulator. The data driven agent converges to the same level as the original $\epsilon$-random RL Agent in Figure \ref{train} within a 1000 episodes, without the need for replays. This promising boost in the learning rate reiterates the need for integrating human intelligence in the form of existing data available on online technical support forums into the learning process of automated learning agents.

\subsubsection{Demo: Interacting with the agent.}
\label{demo}

We offer a demonstration of our system deployed on an Ubuntu shell -- a screen capture of the demonstration is available at the following link:~\url{https://goo.gl/oSmX3d}. First, the user asks for suggestions on how to open Firefox (or asks the agent to open Firefox), and the (trained RL) agent responds by evaluating its $Q$-function with the current state and available actions. Then we make the task a bit harder by uninstalling Firefox and asking again. The agent now responds by activating its root privileges, installs Firefox and opens it, thus demonstrating that it has learned simple dependencies in the Ubuntu OS and can help the user with issues with the same. 


\subsubsection{Work in progress.} 
As an emerging application of AI techniques, there are many avenues of extension. First, we are in the process of expanding the scope of the environment both in terms of the terminal commands available to the agent, as well as the model or representation of the world being learned. Another area of future improvement centers on the data-driven $\epsilon$-random $Q$-learning agent; the current agent is a preliminary exploration into using existing unstructured data to aid the learning process. We are currently looking at augmenting the retrieval mechanism with more advanced 
word-embedding techniques in order to retrieve the most relevant posts from the vast amount of unstructured data available in online technical support forums like \texttt{Ask Ubuntu} and \texttt{Ubuntu Chat Forum}, and as well as structured data available as documentation in manual pages and release notes.

Finally, the current 
work is able to use a basic tabular $Q$-learning approach because the
size of the environment as well as the action space is 
quite limited. As the action and state space sizes increase, our current approach will have to make way for more scalable RL approaches such as function approximation~\cite{sutton1999policy}, and newer approaches that can take into account large action spaces in discrete domains~\cite{dulacdeep}. 





\subsubsection{Acknowledgments}

A significant part of this work was initiated and completed while Tathagata Chakraborti was an intern at IBM's T. J. Watson Research Center.  
The continuation of his work at ASU is supported in part by an IBM Ph.D. Fellowship. Kambhampati's research is supported in part by the ONR grants N000141612892, N00014-13-1-0176, N00014-13-1-0519 and N00014-15-1- 2027.

\bibliographystyle{aaai}
\bibliography{bib}

\end{document}